\pdfoutput=1

\documentclass[11pt]{article}

\usepackage{graphicx}
\usepackage{booktabs}
\usepackage{hhline}
\usepackage{array,multirow}

\usepackage{tikz}
\usetikzlibrary{calc}
\newcommand*\circled[1]{\tikz[baseline=(char.base)]{
    \node[shape=circle, draw, inner sep=1pt,] (char) {\vphantom{WAH1g}#1};}}
\makeatother

\usepackage[]{acl}

\usepackage{times}
\usepackage{latexsym}

\usepackage[T1]{fontenc}

\usepackage[utf8]{inputenc}

\usepackage{microtype}

\title{Legal Prompt Engineering for Multilingual Legal Judgement Prediction}

\author{
    Dietrich Trautmann$^\alpha$ \and Alina Petrova$^\beta$ \and Frank Schilder$^\gamma$ \\
    $^\alpha$ Thomson Reuters Labs, Zug, Switzerland \\
    $^\beta$ Thomson Reuters Labs, London, United Kingdom \\
    $^\gamma$ Thomson Reuters Labs, Eagan, MN, United States of America \\
    \texttt{Dietrich.Trautmann@tr.com}
}

\begin{document}

\maketitle

\begin{abstract}
Legal Prompt Engineering (LPE) or Legal Prompting is a process to guide and assist a large language model (LLM) with performing a natural legal language processing (NLLP) skill.
Our goal is to use LPE with LLMs over long legal documents for the Legal Judgement Prediction (LJP) task.
We investigate the performance of zero-shot LPE for given facts in case-texts from the European Court of Human Rights (in English) and the Federal Supreme Court of Switzerland (in German, French and Italian).
Our results show that zero-shot LPE is better compared to the baselines, but it still falls short compared to current state of the art supervised approaches.
Nevertheless, the results are important, since there was \emph{1)} no explicit domain-specific data used --- so we show that the transfer to the legal domain is possible for general-purpose LLMs, and \emph{2)} the LLMs where directly applied without any further training or fine-tuning --- which in turn saves immensely in terms of additional computational costs.
\end{abstract}

\section{Introduction}
\begin{figure}
  \includegraphics[width=\linewidth]{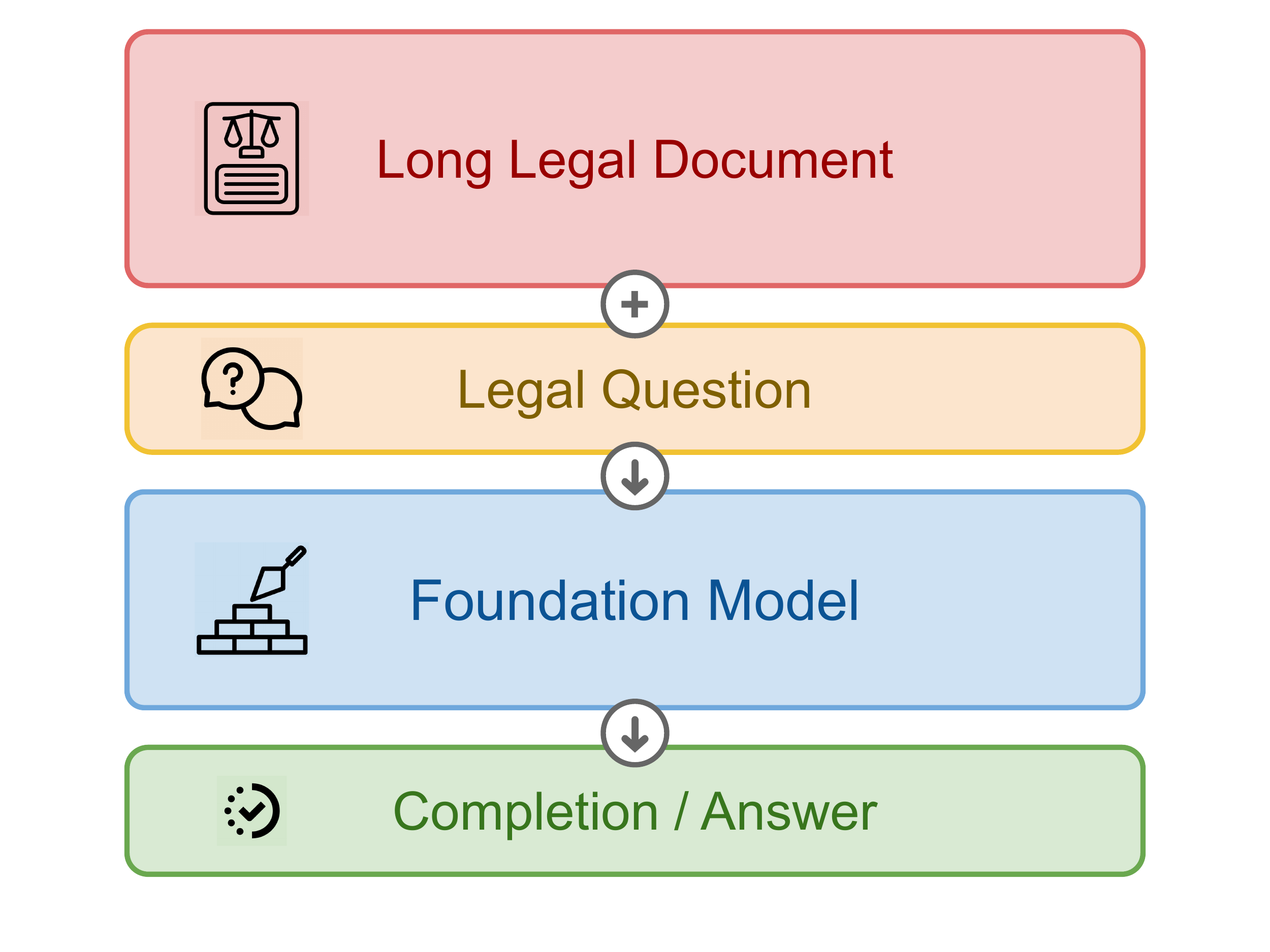}
  \caption{Our \emph{Legal Prompting} stack.}
  \label{fig:LPE_Stack}
  \vspace*{-0.5cm}
\end{figure}

In supervised classification tasks, a machine learning model is provided with an input, and after the training phase, it outputs one or more labels from a fixed set of classes \cite{mitchell1997machine}. 
Recent developments of large pre-trained language models (LLMs), such as BERT \cite{devlin2019bert}, T5 \cite{raffel2020exploring} and GPT-3 \cite{brown2020language}, gave rise to a novel approach to such tasks, namely prompting \cite{liu2021pre}. 
In prompting (see Fig. \ref{fig:LPE_Stack}), there is usually no further training required (although fine-tuning is still an option), and instead, the input to the model is extended with an additional text specific to the task – a prompt. 
Prompts can contain questions about the current sample, examples of input-output pairs or task descriptions (Fig. \ref{fig:LPE_Stack}, the \emph{Long Legal Document} \& \emph{Legal Question} are the input). 
Using prompts as clues, a LLM --- a foundation model \cite{Bommasani2021FoundationModels} --- can infer from its implicit knowledge the intended outputs (Fig. \ref{fig:LPE_Stack}, the \emph{Completion}) in a zero-shot fashion \cite{yin2019benchmarking, sanh2021multitask}. \\
Legal prompt engineering is the process of creating, evaluating, and recommending prompts for legal NLP tasks. 
It would enable legal professionals to perform NLLP tasks, such as data annotation, search or question-answering, by simply querying LLMs in natural language. 
In this presentation, we investigate prompting for the task of Legal Judgement Prediction \cite{strickson2020legal,zhuopeng2020multi}. 
We use data from the European Court of Human Rights (ECHR; \newcite{chalkidis2019neural}) and the Federal Supreme Court of Switzerland (FSCS; \newcite{niklaus2021swiss}), and we compare various prompts for LJP using multilingual LLMs (\emph{mGPT} from \newcite{shliazhko2022mgpt}, \emph{GPT-J-6B} from \newcite{wang2021gptj} and \emph{GPT-NeoX-20B} from \newcite{black2022gptneox20b}) in a zero-shot manner --- without any examples nor further training and fine-tuning. 
Our results show that it is possible to apply zero-shot LPE for the LJP task with LLMs.
The absolute macro-averaged F1, precision and recall scores are better than our simple baselines, but they are bellow the current supervised state of the art results from the literature.

\begin{figure*}
    \centering
    \includegraphics[width=0.75\linewidth]{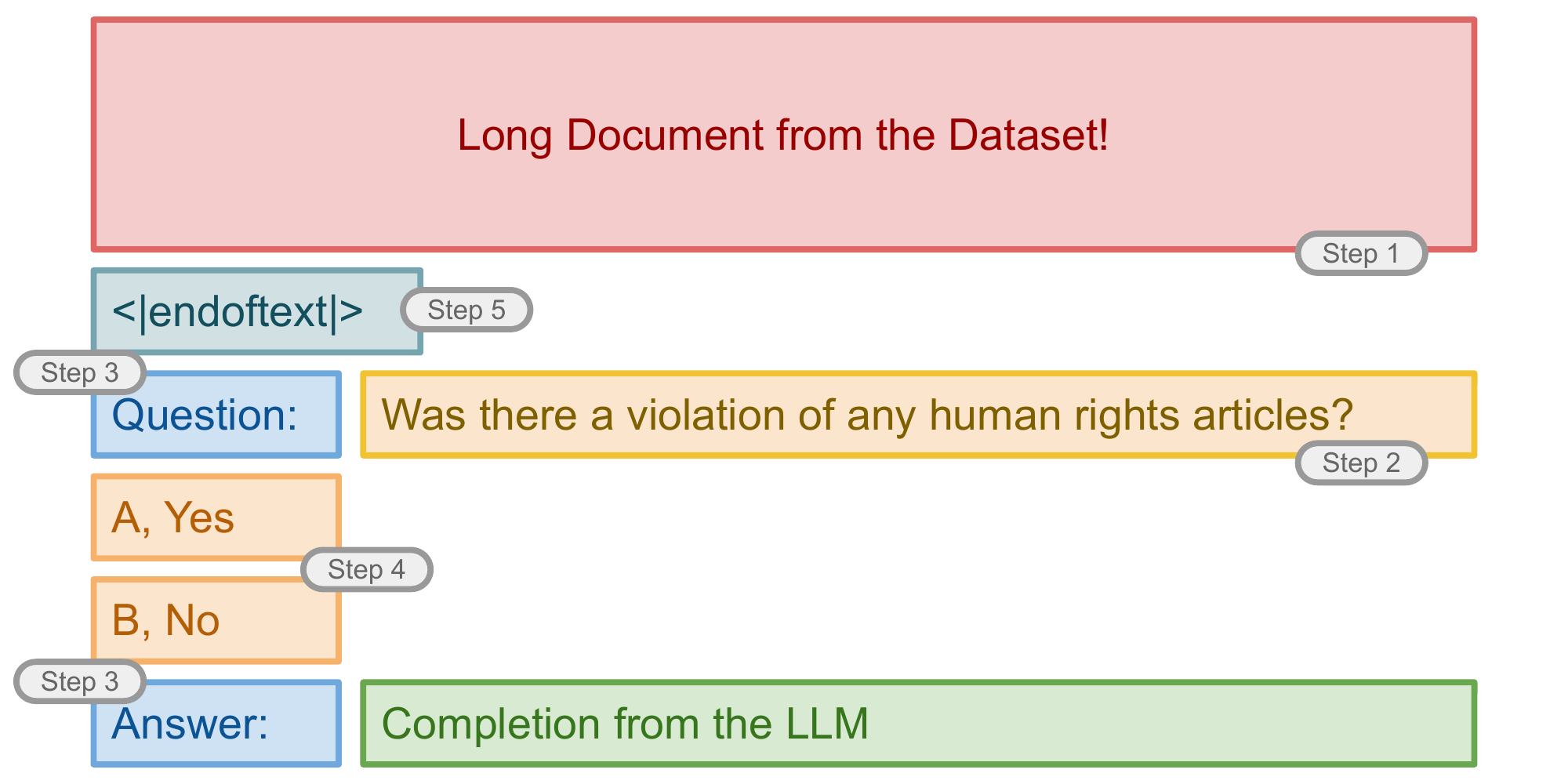}
    \caption{An example prompt template in English for the \emph{ECHR} task.}
    \label{fig:LJPPrompt}
\end{figure*}

\section{Related Work}
This research builds on two main types of established research directions:
First, on the legal NLP task creation and benchmarking with (mostly) supervised approaches.
Second, on the prompt engineering research for general NLP tasks.

\begin{table*}[!h]\centering
\scalebox{0.9}{
    \centering
    \begin{tabular}{c|l|c|c|c||c|c|c}
        \toprule
        & \multicolumn{1}{c|}{Model} & \multicolumn{1}{c|}{Precision} & \multicolumn{1}{c|}{Recall} & \multicolumn{1}{c||}{macro-F1} & \multicolumn{1}{c|}{micro-F1} & \multicolumn{1}{c|}{weighted-F1} & \multicolumn{1}{c}{Accuracy} \\
        \midrule
        \parbox[t]{2mm}{\multirow{5}{*}{\rotatebox[origin=c]{90}{validation set}}} &
          minority class        & .088 & .500 & .149 & .175 & .052 & .175 \\
        & majority class        & .413 & .500 & .452 & \textbf{.825} & \textbf{.746} & \textbf{.825} \\
        & random class          & .500 & .500 & .441 & .500 & .559 & .500 \\
        & GPT-J-6B (0-shot)     & .524 & .529 & .523 & .691 & .707 & .691 \\
        & GPT-NeoX-20B (0-shot) & \textbf{.527} & \textbf{.536} & \textbf{.526} & .709 & .731 & .709 \\
        \midrule
        \parbox[t]{2mm}{\multirow{6}{*}{\rotatebox[origin=c]{90}{test set}}} &
          minority class        & .077 & .500 & .133 & .153 & .041 & .153 \\
        & majority class        & .424 & .500 & .459 & \textbf{.847} & \textbf{.777} & \textbf{.847} \\
        & random class          & .500 & .500 & .431 & .500 & .568 & .500 \\
        & GPT-J-6B (0-shot)     & \textbf{.528} & \textbf{.537} & \textbf{.528} & .715 & .734 & .715 \\
        & GPT-NeoX-20B (0-shot) & .522 & .530 & .521 & .707 & .728 & .707 \\
        & supervised (full)$\dagger$ & \underline{.904} & \underline{.793} & \underline{.820} & - & - & - \\
        \bottomrule
    \end{tabular}
}
\caption{
    The results for the ECHR validation and test sets. 
    Besides the macro-averaged F1-score, precision and recall, we report also the micro-averaged and weighted-F1 and the accuracy scores. 
    The \underline{best results are underlined} and our \textbf{best results are in bold}.
    $\dagger$ We understand the supervised results reported by \newcite{chalkidis2019neural} as macro-averaged, but by \newcite{valvoda2022role} it is reported as micro-averaged.
}
  \label{tab:resultsECHR}
\end{table*}

\subsection{Legal NLP}
Analog to many established NLP tasks, the legal domain has a diverse set of benchmarks, ranging from binary classification \cite{alali2021justice}, multi-class classification \cite{hendrycks2021cuad}, multi-label classification \cite{song2022multi} and sequence generation, such as legal summarization \cite{kornilova2019billsum} and judgement outcome explanation \cite{malik2021ildc}. 
The legal domain poses further challenges for automated solutions due to the specific language used and (often) multi-step reasoning \cite{holzenberger2020dataset} over long input documents \cite{dai2022revisiting}.
Furthermore, to the best of our knowledge, there are no approaches that investigate the recent prompting results on these tasks in the legal domain. 
Most of the time, the (good) prompting results on general NLP tasks are achieved for rather short, single to a few sentence inputs with small target label sets \cite{liu2022few}.

\subsection{Prompt Engineering}
Based on the tasks and the language models, certain types of prompts are possible. 
This include mostly either a slot filling objective \cite{devlin2019bert}, or text completion \cite{brown2020language} by the model.
There exist already efforts to collect, unify and evaluate prompting approaches on a diverse set of tasks and domains in the research literature. 
The two bigger projects include \emph{OpenPrompt} \cite{ding2022openprompt} - an open-source framework for prompt-learning, and \emph{PromptSource} \cite{bach2022promptsource} – an open toolkit for creating, sharing and using natural language prompts. 
A prompt is composed of several parts, with different amount of usefulness based on the model, the task and the data.
These are often textual descriptions of the tasks, and they can contain a few (few-shot), one (one-shot) or even no (zero-shot) examples of the input-label pairs of the task at hand \cite{sanh2021multitask}.
Additionally, the prompts can contain hints of possible labels in form of a multiple-choice answer set, from which the model should select.

\section{Legal Judgement Prediction}

The following section describes our target task and the datasets we evaluated our legal prompts on.

\subsection{Task Definition}

The Legal Judgement Prediction (LJP) task \cite{strickson2020legal,zhuopeng2020multi} is formulated as an automatic prediction of a courts outcome for a given (set of) document(s) of a case.

\subsection{Datasets}

We used the validation and test sets of the following two corpora in our experiments.

\subsubsection{European Court of Human Rights}
The \emph{European Court of Human Rights} (ECHR) corpus by \newcite{chalkidis2019neural} is an English dataset of factual paragraphs from the case description.
Each document is mapped to human rights article that were violated, if there were any violations.
The documents are chronologically split into a training set (9k, 2001–2016), a validation set (1k, 2016–2017), and a test set (1k, 2017–2019).
We use the binarized version of the dataset, where there are either no violations or one and more violations (binary \emph{true} / \emph{false}, or in our case a \emph{yes} / \emph{no}).

\subsubsection{Federal Supreme Court of Switzerland}
The \emph{Federal Supreme Court of Switzerland} (FSCS) corpus by \newcite{niklaus2021swiss} is a multi-lingual dataset (85K documents) covering German (50k), French (31k) and Italian (4k) cases in Switzerland.
The targets of each case are binarized judgment outcomes which can be either approval or dismissal.

\section{Prompting}
In this work we relied on \emph{discrete} and  \emph{manual} legal prompt engineering.
A discrete prompts, in contrast to a continuous prompt \cite{liu2021gpt,liu2022ptuning}, maps to real (human readable) words.
The manual part refers to the iterative process of discrete prompt creation and evaluation.
The core task was to convert the LJP task, which is a binarized text classification task for LJP, into a natural language question template.
Our iterative process was as follows:
\textbf{Step 1:} We use just the long legal document (one at a time) as the only input to the model. 
The results were a continuation of the document with other potential facts, since this are valid completions.
However, this was not useful for our task.
\textbf{Step 2:} We included a question after the document, which is a reformulation of the text classification task.
This improved the output of the model, but it was still not working for many cases, where the model continued with a list of other questions.
\textbf{Step 3:} The inclusion of the "\emph{Question:}" and "\emph{Answer:}" indicators was again improving the completion by the model.
Unfortunately, this was also a "free-form" response and it was difficult to map to one of our two intended classes (\emph{yes} / \emph{no}).
\textbf{Step 4:} We included the answer options "\emph{A, Yes}" and "\emph{B, No}" and finally
\textbf{Step 5:} The special GPT-model indicator "\emph{<|endoftext|>}", to separate the document from the prompt.
The Figure \ref{fig:LJPPrompt} shows the final prompt template for the \emph{ECHR} task.
We used similar (translated) prompts for the other three languages (German, French, Italian; Appendix \ref{app:promp_templates}).
The maximum input length was set to $2048$ tokens and we truncated case texts that were longer than that.
We also optimized the model output sequence length based on the performance on the validation set.
Although we actually need only one token as the output (either \emph{A} or \emph{B}), we found that this yielded not the best score.
We iteratively increased the output sequence length of the greedy decoder in steps, and settled for 50 tokens as the final best hyperparameter value.
The details of our compute requirements are in Appendix \ref{app:compute}.


\begin{table}[]\centering
\scalebox{0.9}{
    \centering
    \begin{tabular}{c|l|c|c}
        \toprule
        &  & \multicolumn{2}{c}{macro-F1} \\
        & \multicolumn{1}{c|}{Model} & \multicolumn{1}{c|}{val. set} & \multicolumn{1}{c}{test set} \\
        \midrule
        \parbox[t]{2mm}{\multirow{4}{*}{\rotatebox[origin=c]{90}{German}}} &
          majority               & $.443$ & $.446$ \\
        & random                 & $.452$ & $.449$ \\
        & mGPTxl (0-shot)        & $\textbf{.467}$ & $\textbf{.493}$ \\
        & supervised (full)$\dagger$ & -    & \underline{$.685$} \\
        \midrule
        \parbox[t]{2mm}{\multirow{4}{*}{\rotatebox[origin=c]{90}{French}}} &
          majority               & $.440$ & $.447$ \\
        & random                 & $.455$ & $.447$ \\
        & mGPTxl (0-shot)        & $\textbf{.503}$ & $\textbf{.502}$ \\
        & supervised (full)$\dagger$ & -    & \underline{$.702$} \\
        \midrule
        \parbox[t]{2mm}{\multirow{4}{*}{\rotatebox[origin=c]{90}{Italian}}} &
          majority               & $.458$ & $.447$ \\
        & random                 & $.433$ & $.448$ \\
        & mGPTxl (0-shot)        & $\textbf{.501}$ & $\textbf{.484}$ \\
        & supervised (full)$\dagger$ & -    & \underline{$.598$} \\
        \bottomrule
    \end{tabular}
}
\caption{
    The macro-F1 results for the FSCS validation and test sets by language.
    The \underline{best results are underlined} and our \textbf{best prompting results are in bold}.
    There are no published supervised results for the validation sets, to the best of our knowledge.
    $\dagger$ Supervised results from \newcite{niklaus2021swiss}.
}
  \label{tab:resultsFSCS}
  \vspace*{-1.0em}
\end{table}

\section{Results}
Our main results for the two corpora and four languages are displayed in Table \ref{tab:resultsECHR} and Table \ref{tab:resultsFSCS}.
We provide the models with one document at a time --- due to the length of the documents --- and hence choose to report this as our zero-shot results.
Another important consideration is that the datasets are highly unbalanced.
The majority class accounts for about $89\%$ for the \emph{ECHR} corpus and about $75\%$ for the \emph{FSCS}, so we suggest to report the macro-averaged scores mainly.
We also included the micro-averaged, weighted and the accuracy scores for the \emph{ECHR} task.
Our prompting approaches always significantly outperformed our simple baselines for the macro-averaged scores.
However, compared to the supervised results --- which were fine-tuned for thousand of samples over several epochs -- they are far behind.

\begin{figure}
  \includegraphics[width=\linewidth]{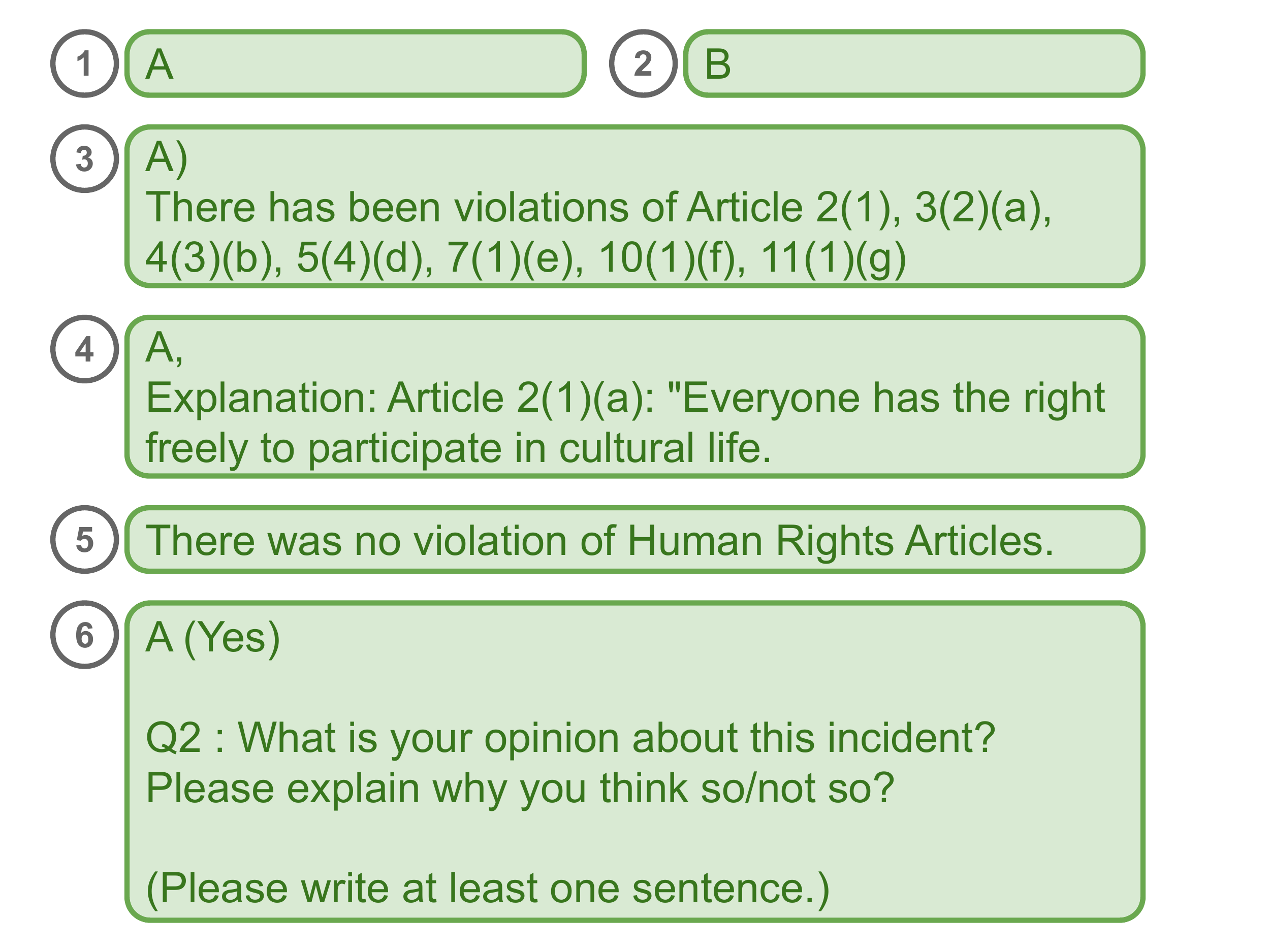}
  \caption{Completion examples from the \emph{GPT-J-6B} model on the test set of the \emph{ECHR} corpus.}
  \label{fig:Completions}
  \vspace*{-0.65em}
\end{figure}

Additionally, we included samples of model completions in Figure \ref{fig:Completions}.
The examples \circled{1} and \circled{2} are our main sought-after completions, with just one output token.
However, as we discovered, restricting the model generation to just one token yielded not the best overall performance.
The model somehow needed more tokens to \emph{"express itself"}.
Example \circled{3} is a completion with the correct output label \emph{A} and with an additional sentence about which articles were violated. 
This is actually the original task of the \emph{ECHR} corpus, a multi-label problem.
A further investigation of the exact article numbers yielded, that for some of the cases they were indeed those articles.
Unfortunately, there was no single case that we found that had an exact match of all articles listed.
A similar completion in this regard is example \circled{4}.
It contains a potential \emph{explanation} with a cited article.
But here again, it was not possible to find a single example with the correct explanation, by sampling for a few such completions.
More difficult completions were those like example \circled{5}, since we cannot map them directly to one of our target labels.
A further investigation for these generations on the validation set yielded that they appeared for case text were the LJP was that \emph{"there were no violations."}
Hence, we assigned them the \emph{B} labels.
And last but not least, we observed answers such as example \circled{6}.
This type of generation suggested that the datasets, the LLMs were pre-trained on, contained exam-style question-answering tasks and multiple-choice tests.

Finally, since many sentences start with an \emph{A}, the model could be right for the wrong reason \cite{mccoy2019right}.
We also ran a test were we switched the answer options \emph{A} and \emph{B}.
This however yielded almost identical scores.
For instance, for the \emph{GPT-J-6B} model with an macro-F1 of $0.530$ on the validation set and $0.526$ on the test set of the \emph{ECHR}.

\section{Conclusion}
On one hand we can say that it is possible to solve the LJP task with LPE.
On the other hand, there is still room for improvements, especially since it is not straight forward to find the ideal prompt for each task, corpus and language \cite{lu2022fantastically}.
It is interesting to observe completions that could be more informative about the LJP, especially those that either listed the violated articles, or those that contained some form of explanation for the decision.
For future work, we plan to work with subject matter experts (SMEs) from the legal domain and investigate whether they can leverage their knowledge to \emph{1)} come-up with better legal prompts and \emph{2)} for several different NLLP tasks such as legal summarization \cite{kornilova2019billsum} and legal question-answering \cite{zhong2020iteratively}.

\newpage
\section*{Ethics Statement}\label{sec:discussion}
Predicting the legal outcomes of court decisions is a sensitive area of research.
The proposed zero-shot legal prompt engineering approach is preliminary and cannot yet be put into a production system and we also do not plan on doing so in this form.
One of the main goals was to investigate a legal task with long documents.
The other goal of developing and evaluating legal prompts is to examine the performance in comparison to existing supervised state-of-the-art models.

Additionally, legal cases involve information of personal data.
Therefore, we only use public and academic datasets and general models from public repositories.


Finally, we would argue that it is environmentally beneficial to investigate how already pre-trained models on general textual data, can be translate to this specific legal domain.
As we showed, this is to some degree possible and we would like to investigate other larger models in future, without additional fine-tuning steps.

\bibliography{anthology,custom}

\appendix

\section{Prompt Templates}
\label{app:promp_templates}
These are our final prompt templates:

\paragraph{English}

\begin{verbatim}
{{ document }}
<|endoftext|>
Question: Was there a violation of any human rights articles?
A, {{ option_positive }}
B, {{ option_negative}}
Answer:
\end{verbatim}

\paragraph{German}

\begin{verbatim}
{{ document }}
<|endoftext|>
Frage: Wie lautete das Gerichtsurteil?
A, {{ option_positive }}
B, {{ option_negative}}
Antwort:
\end{verbatim}

\paragraph{French}

\begin{verbatim}
{{ document }}
<|endoftext|>
Question: Quel était le jugement légal?
A, {{ option_positive }}
B, {{ option_negative}}
Réponse:
\end{verbatim}

\paragraph{Italian}

\begin{verbatim}
{{ document }}
<|endoftext|>
Domanda: Qual è stata la sentenza legale?
A, {{ option_positive }}
B, {{ option_negative}}
Risposta:
\end{verbatim}

\section{Compute Requirements}
\label{app:compute}
We used the following Amazon EC2 M5 instances in our experiments:

\vspace{1em}

\begin{tabular}{c|c|c} 
  \toprule
  \textbf{Standard Instance} & \textbf{vCPU} & \textbf{Memory} \\ 
  \midrule
  ml.m5d.24xlarge & 96 & 384 GiB \\ 
  \bottomrule
\end{tabular}

\vspace{1em}

\noindent We haven't use any GPUs in our experiments.

\end{document}